\documentclass[letterpaper, 10 pt, conference]{ieeeconf}  

\IEEEoverridecommandlockouts                             

\usepackage{graphicx}
\usepackage{booktabs}
\usepackage{amsfonts}
\usepackage{amsmath}
\usepackage{hyperref}
\usepackage{amssymb} 
\usepackage{textcomp} 
\usepackage{marvosym} 

\title{\LARGE \bf
AffordGrasp: In-Context Affordance Reasoning for Open-Vocabulary Task-Oriented Grasping in Clutter 
}

\author{
    Yingbo Tang$^{1,3}$,  
    Shuaike Zhang$^{2,3}$,  
    Xiaoshuai Hao$^{3,\text{\Letter},\dagger}$,  
    Pengwei Wang$^{3}$,  
    Jianlong Wu$^{2}$\\
    Zhongyuan Wang$^{3}$,  
    Shanghang Zhang$^{3,4,\text{\Letter}}$  
    \thanks{$^{1}$Institute of Automation, Chinese Academy of Sciences.}  
    \thanks{$^{2}$Harbin Institute of Technology (Shenzhen).}  
    \thanks{$^{3}$Beijing Academy of Artificial Intelligence.}  
    \thanks{$^{4}$State Key Laboratory of Multimedia Information Processing, School of Computer Science, Peking University.} 
    \thanks{Corresponding authors: \text{\Letter}~Xiaoshuai Hao (xshao@baai.ac.cn), \text{\Letter}~Shanghang Zhang (shanghang@pku.edu.cn).}  
    \thanks{$\dagger$Project leader.}  
}

\begin{document}

\maketitle
\thispagestyle{empty}
\pagestyle{empty}

\begin{abstract}
Inferring the affordance of an object and grasping it in a task-oriented manner is crucial for robots to successfully complete manipulation tasks. 
Affordance indicates where and how to grasp an object by taking its functionality into account, serving as the foundation for effective task-oriented grasping.
 However, current task-oriented methods often depend on extensive training data that is confined to specific tasks and objects, making it difficult to generalize to novel objects and complex scenes.
In this paper, we introduce \textit{\textbf{AffordGrasp}}, a novel open-vocabulary grasping framework that leverages the reasoning capabilities of vision-language models (VLMs) for in-context affordance reasoning. Unlike existing methods that rely on explicit task and object specifications, our approach infers tasks directly from implicit user instructions, enabling more intuitive and seamless human-robot interaction in everyday scenarios.
Building on the reasoning outcomes, our framework identifies task-relevant objects and grounds their part-level affordances using a visual grounding module. This allows us to generate task-oriented grasp poses precisely within the affordance regions of the object, ensuring both functional and context-aware robotic manipulation.
Extensive experiments demonstrate that \textit{\textbf{AffordGrasp}} achieves state-of-the-art performance in both simulation and real-world scenarios, highlighting the effectiveness of our method. We believe our approach advances robotic manipulation techniques and contributes to the broader field of embodied AI. 
Project website: \url{https://eqcy.github.io/affordgrasp/}.

\end{abstract}

\section{INTRODUCTION}
For robots assisting in daily tasks, understanding object affordances and selecting grasp areas and poses are vital for effective execution. For example, when pouring water, a robot must identify a cup’s handle as the graspable region to ensure success. This highlights the need for task-oriented grasping based on affordances, akin to how humans intuitively interact with objects—like holding a spoon or racket handle—and generalize to new objects.
However, enabling robots to replicate this ability is challenging, particularly in accurately identifying affordances without extensive task-specific training. Overcoming this is key to developing robots that adapt to diverse real-world scenarios.

\begin{figure}[h]
\centering
\includegraphics[scale=0.60]{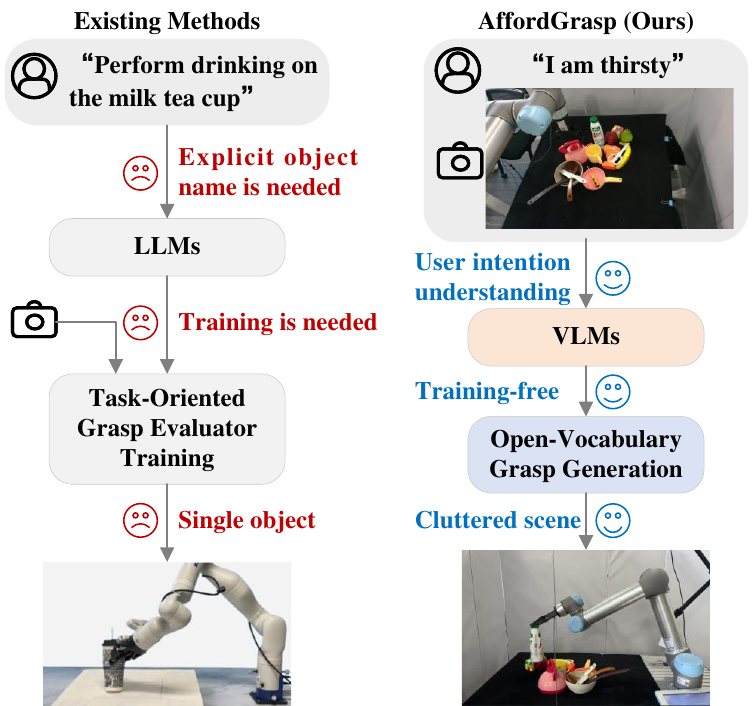}
\caption{Compared to existing task-oriented grasping methods, our approach offers three key advantages: (1) leveraging VLMs for affordance reasoning, enabling better understanding of user intentions from implicit language instructions and visual scene observations; (2) a training-free pipeline, eliminating the need for annotated data to train task-oriented grasp evaluators; and (3) the ability to handle cluttered scenes, rather than being limited to single-object grasping.}
\label{figure_motivation}
\vspace{-0.5em}
\end{figure}

Existing methods \cite{bahl2023affordances,ju2024robo,kuang2024ram,dong2024rtagrasp,tang2023graspgpt,tang2024foundationgrasp} transfer affordances \cite{bahl2023affordances,ju2024robo,kuang2024ram} or task-oriented grasps \cite{dong2024rtagrasp,tang2023graspgpt,tang2024foundationgrasp} from training data to test objects by constructing affordance memories or learning task-specific skills. These approaches retrieve semantically and geometrically similar objects to transfer knowledge from seen to unseen instances. However, their reliance on time-consuming retrieval processes and limited affordance memory scales hinders their effectiveness in open-vocabulary scenarios with novel tasks and objects.

Recent research shows growing interest in open-vocabulary grasping \cite{li2024ovgnet,zheng2024gaussiangrasper,bhat2024hifi,tziafas2024towards}. These methods generate grasp poses by jointly training vision-language grasping frameworks \cite{li2024ovgnet,zheng2024gaussiangrasper,bhat2024hifi} or leveraging the visual grounding abilities of VLMs \cite{tziafas2024towards}. However, they focus solely on grasping entire objects, neglecting task-specific affordances. For finer-grained grasping, \cite{van2024open} uses open-vocabulary object detection \cite{liu2024grounding} and part segmentation \cite{sun2023going} to perform part-level grasps based on user's part prompts. Other works \cite{mirjalili2024lan,tong2024oval} employ Large Language Models (LLMs) to reason about object parts or affordances from language instructions, localizing regions via VLMs. However, existing methods often require explicit object names in prompts and struggle with implicit instructions (\textit{e.g.}, ``I am thirsty'') without visual scene context, lacking the ability to interpret tasks and object functionalities in complex open-vocabulary environments.

Although existing task-oriented or open-vocabulary grasping methods perform well in some scenarios, inferring appropriate open-vocabulary task-oriented grasps in cluttered environments remains challenging. We summarize the key difficulties as follows: \textit{\textbf{(1) Ambiguity in user instructions.}} Existing methods often require explicit object names and tasks, lacking flexibility. In practice, user instructions frequently contain implicit goals, demanding deeper semantic understanding. \textit{\textbf{(2) Training dependency.}} Some methods \cite{tang2023graspgpt,tang2024foundationgrasp} use LLMs to infer object properties and build task-object similarity, yet still rely on annotated data to train task-oriented grasp evaluators for generating desired poses. \textit{\textbf{(3) Cluttered scenes.}} Current task-oriented grasping methods typically focus on single-object scenarios. Cluttered environments with multiple objects increase the complexity of identifying target objects, posing additional challenges.

To address the aforementioned challenges, we propose \textit{\textbf{AffordGrasp}}, a novel affordance reasoning framework for open-vocabulary task-oriented grasping. As illustrated in Fig. \ref{figure_motivation}, our method offers three key advantages over existing approaches. First, we tackle ambiguous user instructions by leveraging Vision-Language Models (VLMs) to reason about affordances from both language and visual inputs. By utilizing the inherent reasoning capabilities of VLMs, we establish relationships among tasks, objects, and affordances in complex scenes. Second, based on the affordance reasoning results, the target object and its visual affordance are grounded using VLMs for grasp generation, eliminating the need for additional training data. Finally, with a deep understanding of tasks and affordances, our method outperforms existing approaches in cluttered scenes, enabling robust and context-aware grasping.
The main contributions of our work are as follows:

\begin{itemize}
\item We introduce AffordGrasp, a novel framework for open-vocabulary task-oriented grasping in cluttered scenes.

\item Specifically, we propose an in-context affordance reasoning module that leverages the reasoning capabilities of Vision-Language Models (VLMs). This module extracts explicit tasks from implicit user instructions and infers object affordances from visual observations, significantly enhancing the understanding of tasks, objects, and their affordances through the integration of language and visual inputs.

\item Importantly, our method is entirely open-vocabulary, requiring only arbitrary user instructions and RGB-D images. It functions without the need for additional training or fine-tuning, which makes it highly flexible and scalable.

\item We validate the effectiveness of our approach through extensive experiments in both simulation and real-world robotic scenarios, demonstrating its robustness and practical applicability.
\end{itemize}

\section{RELATED WORK}
\subsection{Visual Affordance Grounding}

Visual Affordance Grounding involves identifying and localizing specific areas of an object in an image that enable potential interactions. Researchers have explored various approaches to learn affordances, including annotated images \cite{luo2022learning}, demonstration videos \cite{chen2023affordance,bahl2023affordances}, and cross-modal inputs \cite{li2024laso}. Methods like Robo-ABC \cite{ju2024robo} and RAM \cite{kuang2024ram} employ semantic correspondence and retrieval-based paradigms to generalize affordances across objects and domains, utilizing diverse data sources for zero-shot transfer. Additionally, GLOVER \cite{ma2024glover} fine-tunes a multi-modal LLM to predict visual affordances. However, these approaches often require varying degrees of data training, limiting their applicability in fully open-vocabulary scenarios.

\subsection{Task-Oriented Grasping}
Task-Oriented Grasping focuses on grasping specific object parts based on task constraints, enabling robots to perform subsequent manipulations. Traditional approaches rely on training with large-scale annotated datasets \cite{murali2021same,song2023learning}. To enhance generalization to novel objects, methods like GraspGPT \cite{tang2023graspgpt} and FoundationGrasp \cite{tang2024foundationgrasp} leverage the open-ended semantic knowledge of LLMs to bridge connections between dataset objects and novel ones, though they still depend on annotated datasets for training. LERF-TOGO \cite{lerftogo2023} constructs a Language-Embedded Radiance Field (LERF) \cite{kerr2023lerf} and generates 3D relevancy heatmaps using DINO features \cite{caron2021emerging} for zero-shot task-oriented grasping. However, it requires time-consuming multi-view image capture for 3D rendering and explicit object part specifications. ShapeGrasp \cite{li2024shapegrasp} employs geometric decomposition, including attributes and spatial relationships, to grasp novel objects. Despite these advancements, existing methods primarily focus on single-object grasping, limiting their effectiveness in cluttered environments.

\subsection{Robotic Grasping with LLMs and VLMs}
The integration of LLMs and VLMs into robotic grasping has recently garnered significant interest. Approaches like LAN-Grasp \cite{mirjalili2024lan} leverage LLMs to identify graspable object parts, while OVAL-Prompt \cite{tong2024oval} formulates this as affordance grounding. However, these methods rely exclusively on LLMs to infer grasping positions from language instructions, lacking integration with visual context. Consequently, they depend on explicit instructions and struggle in cluttered environments without advanced reasoning. Recent advancements \cite{jin2024reasoning,xu2024rt,qian2024thinkgrasp} emphasize the strong contextual reasoning capabilities of VLMs. For example, ThinkGrasp \cite{qian2024thinkgrasp} introduces a vision-language grasping system for cluttered scenes. While ThinkGrasp represents progress in clutter handling, it focuses on object-level grasping and overlooks task constraints and affordances, which are essential for fine-grained, task-oriented grasping.

\section{METHOD}
\label{method}

\begin{figure*}[ht] 
\centering
\includegraphics[scale=0.66]{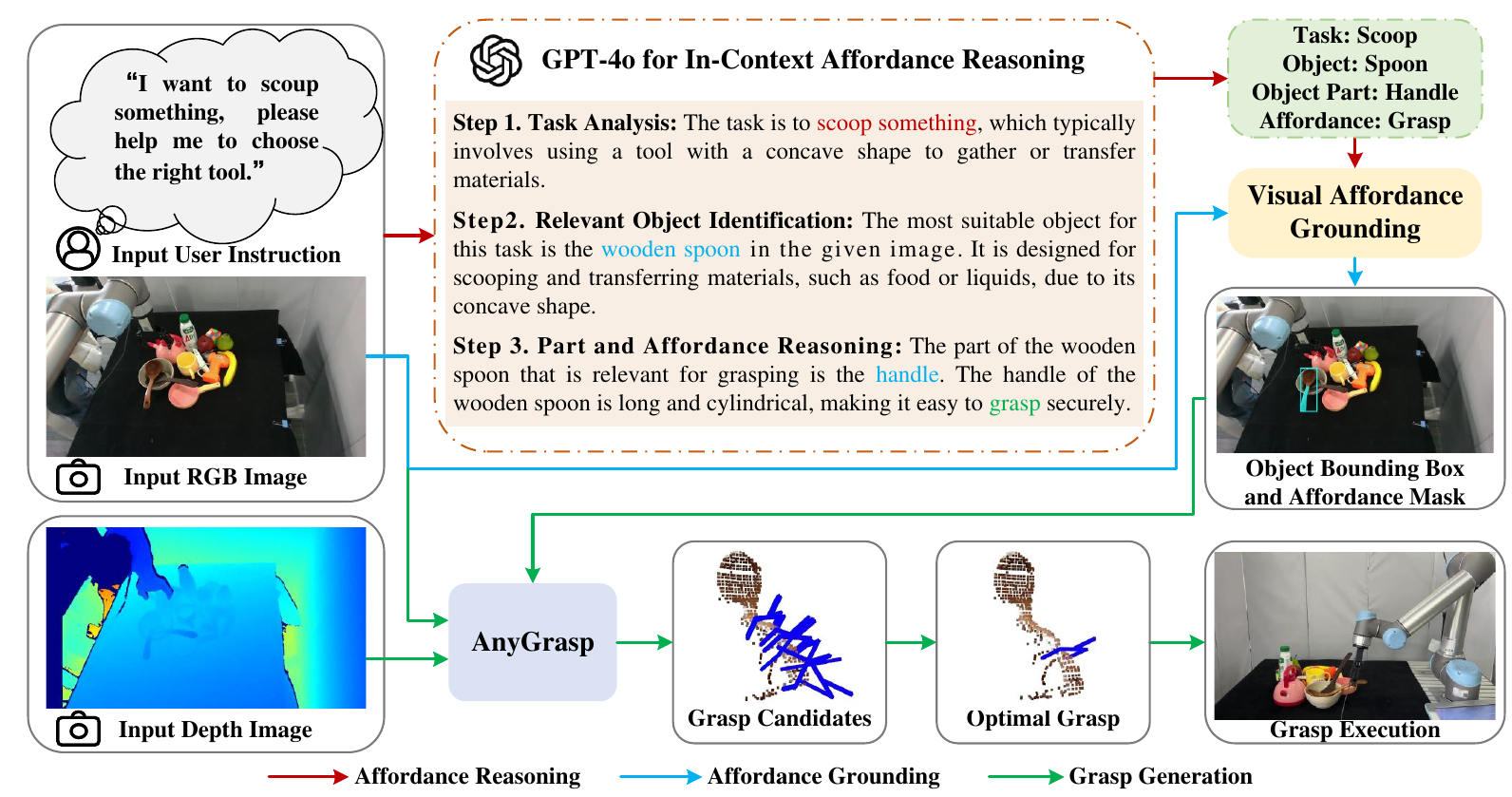}
\caption{
\textbf{Overall Framework of AffordGrasp}.
The framework processes user instructions and RGB-D scene observations to achieve open-vocabulary task-oriented grasping in clutter. We leverage GPT-4o \cite{hurst2024gpt} for in-context affordance reasoning, decomposing the process into three steps:
(1) Extracting the task goal and functional requirements from implicit user instructions (\textit{e.g.}, ``I want to scoop something'').
(2) Identifying the most task-relevant object in the RGB image (\textit{e.g.}, a wooden spoon).
(3) Decomposing the object into functional parts and selecting the optimal graspable part (\textit{e.g.}, the handle) based on its affordances.
Based on the reasoning results, a visual affordance grounding module grounds the inferred object and part affordances into pixel-level masks. With the affordance mask and RGB-D images, we employ AnyGrasp \cite{10167687} to generate task-oriented 6D grasp poses on the target part.
}
\label{figure_framework}
\vspace{-0.5em}
\end{figure*}

In this section, we introduce \textit{\textbf{AffordGrasp}}, a novel framework for open-vocabulary task-oriented grasping. As illustrated in Fig. \ref{figure_framework}, our method begins by interpreting user instructions through a Vision-Language Model (VLM), which decomposes the task into three key steps:  analyzing the task goal, identifying the most relevant object, and reasoning about the optimal graspable part based on affordances. This reasoning process is grounded into pixel-level masks using a visual grounding module, which segments the object and its functional parts in the RGB image. Leveraging the RGB-D image and affordance mask, we employ AnyGrasp \cite{10167687} to generate 6D grasp poses, aligning the affordance center to select the optimal grasp. By integrating language-guided reasoning, visual grounding, and geometry-aware grasp synthesis, \textit{\textbf{AffordGrasp}} enables robust and interpretable task-oriented grasping in cluttered environments, effectively bridging high-level task semantics with low-level robotic execution.

\begin{figure*}[ht] 
\centering
\includegraphics[scale=0.60]{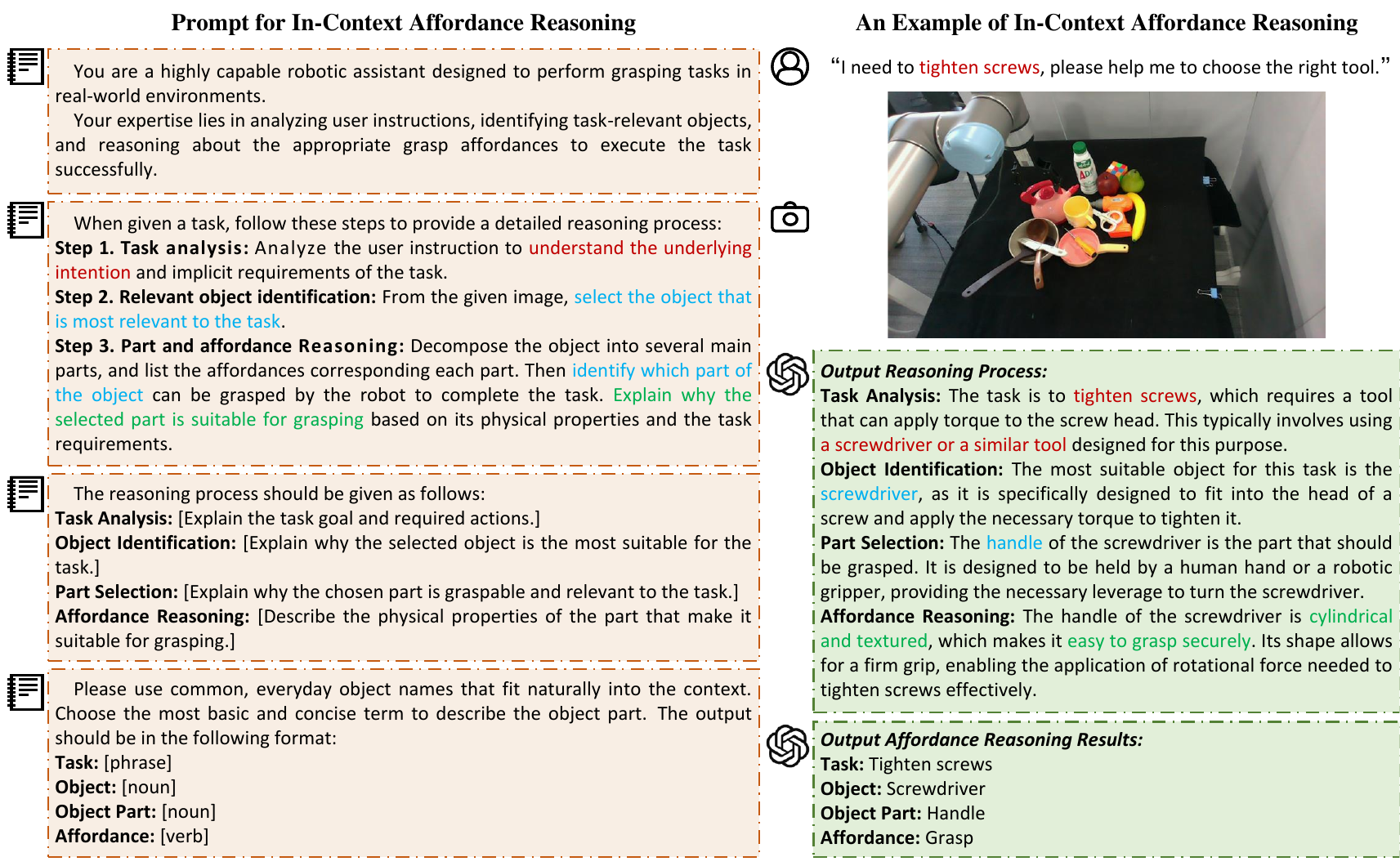}
\caption{
The prompt for in-context affordance reasoning is designed, and an example is provided for illustration.
}
\label{figure_prompt}
\end{figure*}

\subsection{In-Context Affordance Reasoning}
Inferring affordances for task-oriented grasping in cluttered environments is challenging, as it requires interpreting user instructions, identifying task-relevant objects, and reasoning about functional parts and affordances. To address this, we propose an in-context affordance reasoning module powered by Vision-Language Models (VLMs), which decomposes the problem into three key steps: (1) task analysis, (2) relevant object identification, and (3) part and affordance reasoning. Specifically, we leverage the GPT-4o \cite{hurst2024gpt} model to analyze the underlying intention and implicit requirements of the task from user instructions, and then establish relationships among tasks, objects, and affordances in an in-context manner using visual observations.

\textbf{Task Analysis.}  
Given a language instruction $\mathcal{L}$ and an RGB image of the scene $\mathcal{I}$, we first analyze the user instruction using GPT-4o to extract the explicit task $\mathcal{T}$.

\textbf{Relevant Object Identification.}  
Based on the requirements of the explicit task $\mathcal{T}$, the object $\mathcal{O}$ most relevant to the task is identified in the image.

\textbf{Part and Affordance Reasoning.}  
The selected object $\mathcal{O}$ is decomposed into functional parts $\mathcal{P} = \{p^1, p^2, \dots, p^n\}$, and the corresponding affordances $\mathcal{A} = \{a^1, a^2, \dots, a^n\}$ of each part are inferred. The optimal part $p^*$ and its affordance $a^*$ are selected based on their suitability for the task. The entire reasoning process can be formulated as follows:
\begin{equation}
    \mathcal{T},\: \mathcal{O},\: p^*,\: a^* = \text{GPT-4o}(\mathcal{L},\: \mathcal{I}).
\end{equation}

Fig.~\ref{figure_prompt} illustrates the prompt for the in-context affordance reasoning module, including an example of input and output. This prompt guides the Vision-Language Model (VLM) through a step-by-step reasoning process: (1) understanding user intent, (2) identifying the most relevant object, and (3) locating the optimal graspable part based on affordances. This method enhances interpretability and ensures task-oriented grasp generation.
Our approach effectively handles complex scenarios with multiple objects, enabling the robot to prioritize relevant items while ignoring distractions. The reasoning output is structured as task, object, object part, and affordance, serving as input for subsequent visual affordance grounding and grasp pose generation.

\subsection{Visual Affordance Grounding}
To ground the object part and corresponding affordance on the task-relevant object, we propose a visual affordance grounding module. Unlike traditional methods trained on closed-set semantic concepts \cite{luo2022learning,chen2023affordance}, our approach leverages the open-vocabulary part segmentation method VLPart \cite{sun2023going}, making it open-ended. The module takes as input the RGB image $\mathcal{I}$ and the affordance reasoning results $\{\mathcal{O},\: p^*,\: a^*\}$, and outputs an object bounding box with an affordance mask on the object part.

The process is divided into two steps:
\begin{enumerate}
    \item \textbf{Object Localization:} VLPart locates the object bounding box $\mathcal{B_O}$ and generates a masked image $\mathcal{M_{B_O}}$:
    \begin{equation}
        \mathcal{B_O} = \text{VLPart}(\mathcal{I},\: \mathcal{O}),
    \end{equation}
    \begin{equation}
        \mathcal{M_{B_O}}(i,\: j) = 
        \begin{cases}
            \mathcal{I}(i,\: j) & \text{if } (i,\: j) \in \mathcal{B_O}, \\
            0 & \text{if } (i,\: j) \notin \mathcal{B_O}.
        \end{cases}
    \end{equation}

    \item \textbf{Affordance Mask Prediction:} The affordance mask $\mathcal{M}_{p^*}$ is predicted on the masked image using the optimal part $p^*$ and affordance $a^*$:
    \begin{equation}
        \mathcal{M}_{p^*} = \text{VLPart}(\mathcal{M_{B_O}},\: p^*,\: a^*).
    \end{equation}
\end{enumerate}

This two-step approach effectively eliminates interference from other objects in cluttered  scenes.

\subsection{Grasp Pose Generation}
We employ AnyGrasp \cite{10167687} to generate 6D grasp poses based on visual affordance. First, the depth image is converted into a partial-view point cloud using camera intrinsics. The affordance mask from the visual affordance grounding module is then used to filter the point cloud, enabling task-oriented grasp generation. This strategy constrains the grasp generation process and effectively avoids interference from surrounding objects.

\begin{table*}[ht]
\caption{Simulation Results for Various Methods in Single Object Grasping.}
\label{table_simulation_1}
\begin{center}
\begin{tabular}{l|c c c c c c c|c}
\toprule
Methods     & Cup              & Spoon             & Hammer               & Bowl          & Screwdriver      & Scissors              & Wine Glass    & Average GSR
\\
\midrule
GraspNet \cite{fang2020graspnet}    & 0.68            & 0.42              & 0.60                 & 0.84          & 0.84             & 0.14                  & 0.44         & 0.57
\\
ThinkGrasp \cite{qian2024thinkgrasp}   & 0.70            & 0.92              & 0.86                 & \textbf{0.90} & 0.94             & 0.48                  & 0.46         & 0.75
\\
\textbf{AffordGrasp~(Ours)}        & \textbf{0.92}    & \textbf{0.98}     & \textbf{0.88}        & 0.88          & \textbf{1.00}    & \textbf{0.62}         & \textbf{0.66} & \textbf{0.85}
\\
\bottomrule
\end{tabular}
\end{center}
\end{table*}

\begin{table*}[ht]
\caption{Simulation Results of Various Methods for Grasping in Clutter.}
\label{table_simulation_2}
\begin{center}
\begin{tabular}{l|c c c c c c c|c}
\toprule
Methods     & Cup              & Spoon             & Hammer               & Bowl            & Screwdriver      & Scissors            & Wine Glass    & Average GSR
\\
\midrule
ThinkGrasp \cite{qian2024thinkgrasp}  & 0.70             & 0.88              & 0.76                 & \textbf{0.96}   & 0.16             & 0.30                & 0.00         & 0.54
\\
\textbf{AffordGrasp~(Ours)}         & \textbf{0.84}    & \textbf{0.92}     & \textbf{0.90}        & 0.94            & \textbf{0.76}    & \textbf{0.50}       & \textbf{0.52} & \textbf{0.77}
\\
\bottomrule
\end{tabular}
\end{center}
\end{table*}

\begin{figure*}[ht] 
\centering
\includegraphics[scale=0.60]{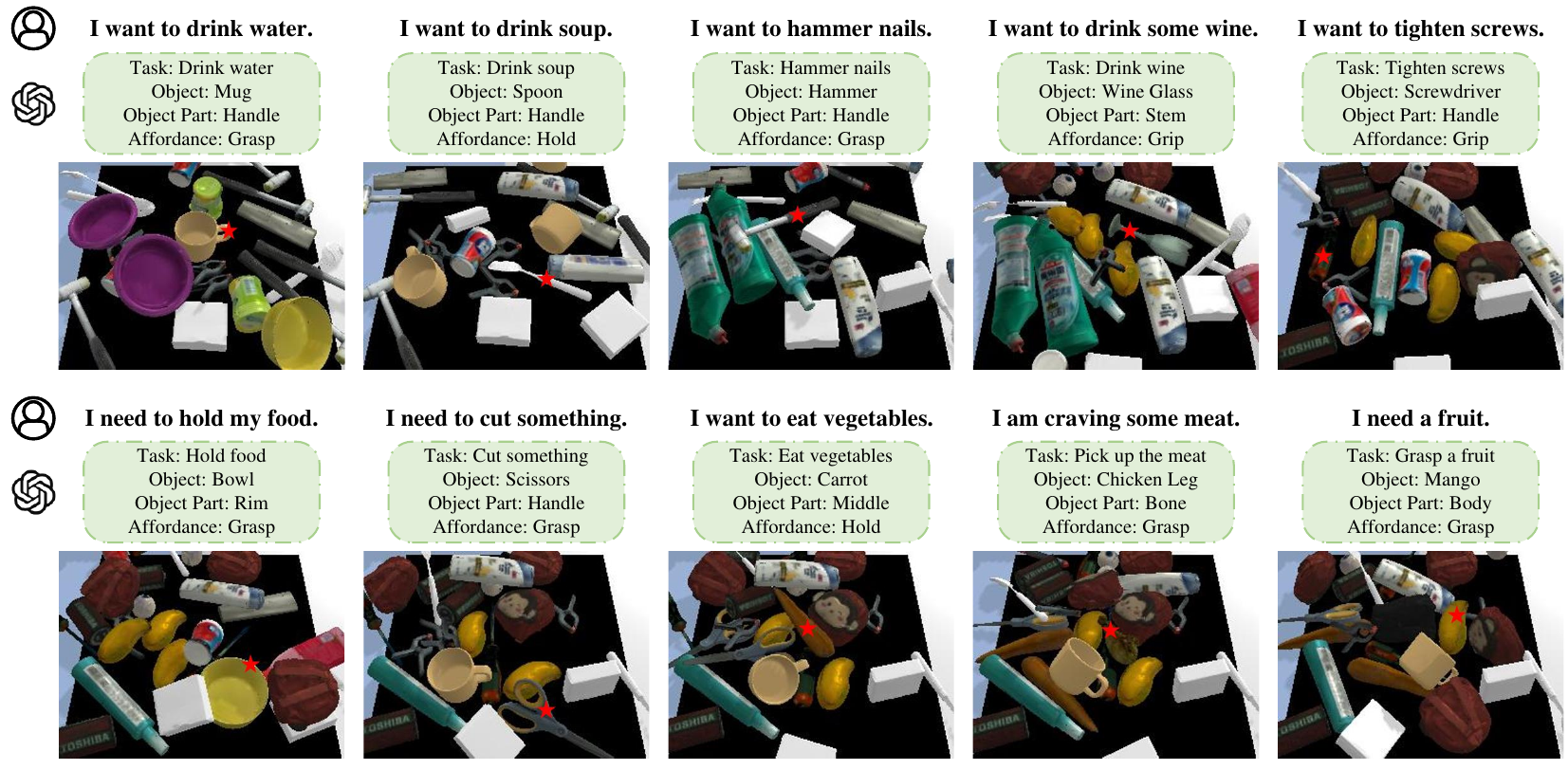}
\caption{\textbf{Simulation Cases of Grasping in Clutter}: The affordances of the target object are indicated with red stars.}
\label{figure_simulation}
\end{figure*}

The point cloud of the object part with grasp affordance is processed by AnyGrasp to generate grasp candidates. AnyGrasp is a 6D grasp generation model trained on a large-scale dataset, which produces grasp poses from partial-view point cloud. Each grasp pose $g$ is represented as:
\begin{equation}
    g = [\mathbf{R}, \mathbf{t}, w],
\end{equation}
where $\mathbf{R} \in \mathbb{R}^{3 \times 3}$ is the rotation matrix, $\mathbf{t} \in \mathbb{R}^{3 \times 1}$ is the translation vector, and $w \in \mathbb{R}$ is the minimum gripper width. The optimal grasp pose $g^*$ is selected from candidates $\mathcal{G} = \{g_1, g_2, \dots, g_n\}$ using:
\begin{equation}
    g^* = \arg \max_{g \in \mathcal{G}} \left( \frac{score(g)}{\| \mathbf{t}(g) - \mathbf{c} \|_2} \right),
\end{equation}
where $score(g)$ and $\mathbf{t}(g)$ are the confidence score and translation vector of $g$, respectively. This prioritizes grasps closer to the affordance mask centroid $\mathbf{c} = [x, y, z]$ with higher confidence scores, ensuring alignment with the affordance's geometric center for improved stability.

\section{EXPERIMENT}

\subsection{Experiment Setup}

\textbf{Implementation Details.} 
All experiments are conducted on a workstation with a 24GB GeForce RTX 4090 GPU. The simulation environment is built in PyBullet \cite{coumans2016pybullet}, featuring a UR5 arm with a ROBOTIQ-85 gripper and an Intel RealSense L515 camera. Raw images are resized to $224 \times 224$ pixels for segmentation. Our real-world setup includes a UR5 arm with an RS-485 gripper and an Intel RealSense L515 camera, calibrated in an eye-to-hand configuration. RGB-D images are captured at $1280 \times 720$ resolution. GraspNet \cite{fang2020graspnet} is used for grasp pose generation in simulations, while AnyGrasp \cite{10167687} is employed for real-world experiments.

\textbf{Baseline Methods.} 
We compare our method with GraspNet \cite{fang2020graspnet} and AnyGrasp \cite{10167687}, both 6D grasping methods trained on large-scale real-world datasets; RAM \cite{kuang2024ram}, which generates 2D affordance maps and lifts them to 3D for manipulation; and ThinkGrasp \cite{qian2024thinkgrasp}, a vision-language grasping system leveraging GPT-4o for reasoning in cluttered environments.

\textbf{Evaluation Metrics.} 
We use Grasp Success Rate (GSR), calculated as the percentage of successful grasps relative to total grasp executions, to evaluate performance in both simulation and real-world experiments.

\begin{table*}[ht]
\caption{Comparison Results of Various Methods in Real-World Experiments.}
\label{table_realworld}
\begin{center}
\begin{tabular}{l|c c c c c c c c|c}
\toprule
Methods     & Bottle    & Mug    & Spatula    & Spoon    & Knife    & Pan    & Kettle    & Screwdriver    & Average GSR
\\
\midrule
RAM \cite{kuang2024ram}         & 6/10    & 1/10    & 0/10    & 0/10    & 0/10    & 0/10    & 0/10    & 0/10    & 0.14
\\
AnyGrasp \cite{10167687}    & \textbf{8/10}    & 3/10    & 7/10    & 6/10    & \textbf{10/10}    & 7/10    & 2/10    & 6/10    & 0.61
\\
ThinkGrasp \cite{qian2024thinkgrasp} & \textbf{8/10}    & 6/10    & 5/10    & 7/10    & 9/10    & 7/10    & \textbf{7/10}    & \textbf{9/10}    & 0.73
\\
\textbf{AffordGrasp~(Ours)}       & \textbf{8/10}    & \textbf{8/10}    & \textbf{9/10}    & \textbf{8/10}    & \textbf{10/10}    & \textbf{9/10}    & 6/10    & 8/10    & \textbf{0.83}
\\
\bottomrule
\end{tabular}
\end{center}
\end{table*}

\begin{figure*}[ht] 
\centering
\includegraphics[scale=0.60]{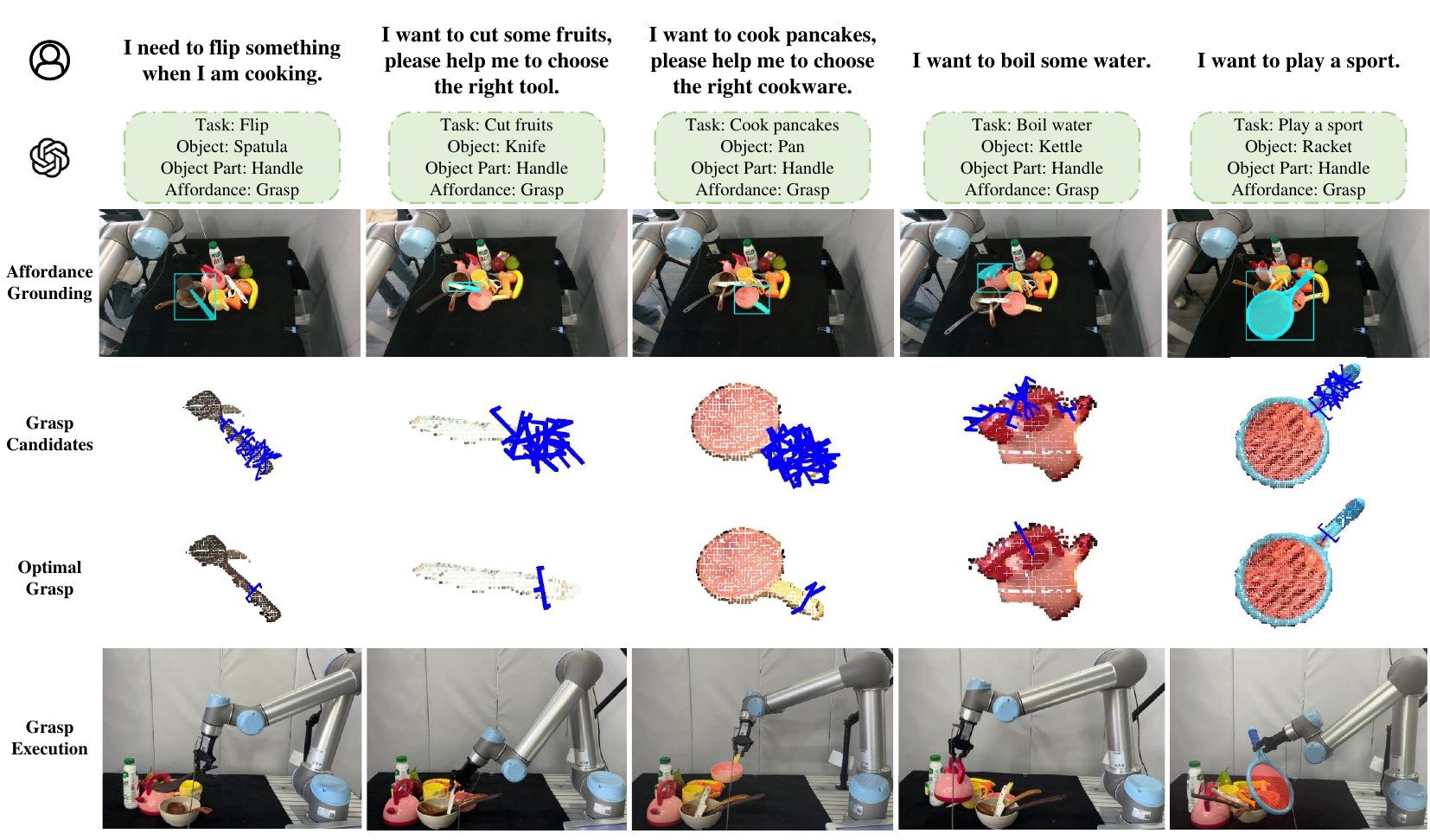}
\caption{\textbf{Real-world Examples of Grasping in Clutter:} The visualization of affordance grounding and task-oriented grasp generation are provided.}
\label{figure_realworld}
\end{figure*}

\subsection{Simulation Results}
We compare our method with GraspNet \cite{fang2020graspnet} and ThinkGrasp \cite{qian2024thinkgrasp} in simulation, using GraspNet for grasp generation. Experiments involve two tasks: \textbf{Grasping Single Object} and \textbf{Grasping in Clutter}. Each case includes 50 runs with identical scene settings for all methods.

\textbf{Grasping Single Object} involves placing objects in isolation on a table. Results in Tab. \ref{table_simulation_1} show our method achieves an average grasp success rate (GSR) of 0.85, outperforming GraspNet and ThinkGrasp. For simple objects (\textit{e.g.}, spoon, hammer, bowl), our method matches ThinkGrasp's performance. For challenging objects like scissors and wine glass, our method achieves GSRs of 0.62 and 0.66, surpassing GraspNet by 0.48 and 0.22, and ThinkGrasp by 0.14 and 0.20, respectively. This demonstrates our method's adaptability to complex geometries.

\textbf{Grasping in Clutter}  involves randomly arranged objects, with target affordances marked by red stars (Fig. \ref{figure_simulation}). Quantitative results in Tab. \ref{table_simulation_2} show our method achieves an average GSR of 0.77, significantly outperforming ThinkGrasp's 0.54. Our method excels with cups, spoons, hammers, and screwdrivers. For screwdrivers, ThinkGrasp obtains only 0.16 GSR, while our method achieves 0.76. ThinkGrasp also struggles with wine glasses, as it prioritizes the most salient object, leading to failures even when the target is visible.

\subsection{Real-world Results}
We evaluate our method on a diverse set of objects in real-world task-oriented grasping in clutter. Results in Tab. \ref{table_realworld} show our method achieves the highest average grasp success rate (GSR) of 0.83, outperforming RAM, AnyGrasp, and ThinkGrasp. For simple objects (\textit{e.g.}, bottle), all methods achieve high GSRs. However, RAM struggles with complex geometries due to limited affordance memory, while ThinkGrasp shows competitive performance on specific objects like kettles and screwdrivers. Our method consistently excels across a broader range of objects, achieving GSRs of 0.80, 0.90, and 0.90 for mugs, spatulas, and pans, respectively, demonstrating robustness across varying shapes and functionalities.

Visualizations of affordance grounding and task-oriented grasp generation are shown in Fig. \ref{figure_realworld}. Relevant objects and affordance masks are labeled with cyan bounding boxes and masks in RGB images. Grasp poses are generated on point clouds corresponding to affordance masks and visualized on the object point cloud. As shown in Fig. \ref{figure_realworld}, grasp poses are densely distributed on object parts identified with grasp affordances by GPT-4o reasoning. The optimal grasp pose is selected using the filtering strategy detailed in Sec.~\ref{method} C for robotic grasp execution.

\subsection{Case Study}

\begin{figure}[ht]
\centering
\includegraphics[scale=0.5]{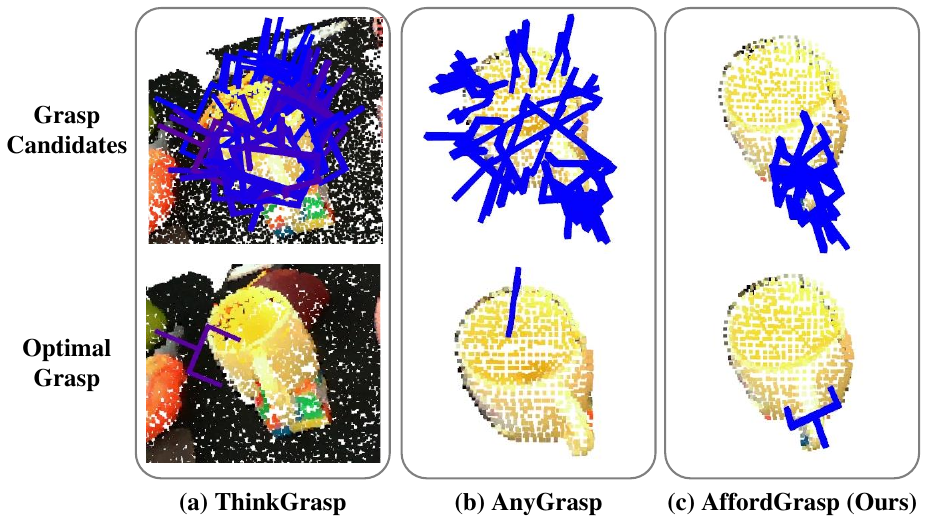}
\vspace{-8pt}
\caption{The case study of mug in real-world experiments.}
\label{figure_case}
\vspace{-0.5em}
\end{figure}

To demonstrate the advantages of our method, we present a case study in Fig.~\ref{figure_case}. The first row shows grasp pose candidates, while the second row depicts the optimal grasp poses selected by different methods. ThinkGrasp selects grasps near the center of the cropped image due to its $3 \times 3$ grid strategy, which divides the image into numbered regions (1: top-left, 9: bottom-right). However, this often fails for objects with curved rims (\textit{e.g.}, mugs), as center-based grasps lack stability due to insufficient contact points or improper force distribution. Similarly, AnyGrasp generates arbitrary grasps without affordance reasoning, leading to unreliable task-oriented grasping (Fig.~\ref{figure_case}(b) and (c)). In contrast, our method integrates affordance reasoning and fine-grained grasp generation, producing precise and stable grasps. This highlights the importance of affordance reasoning and demonstrates the effectiveness of our method in real-world applications.

\section{CONCLUSIONS}
This paper introduces \textit{\textbf{AffordGrasp}}, a novel open-vocabulary task-oriented grasping framework that leverages the reasoning capabilities of VLMs to infer object affordances and generate task-specific grasps in cluttered environments. By integrating in-context affordance reasoning and visual affordance grounding, our method effectively interprets user instructions, identifies task-relevant objects, and grounds part-level affordances for precise grasp generation. Extensive experiments in simulation and real-world scenarios demonstrate the effectiveness of our approach, outperforming existing methods in handling novel objects and complex scenes without requiring additional training data.

\addtolength{\textheight}{-12cm}

\addtolength{\textheight}{12cm} 
\bibliographystyle{IEEEtran}
\bibliography{references}

\begin{thebibliography}{10}
\providecommand{\url}[1]{#1}
\csname url@samestyle\endcsname
\providecommand{\newblock}{\relax}
\providecommand{\bibinfo}[2]{#2}
\providecommand{\BIBentrySTDinterwordspacing}{\spaceskip=0pt\relax}
\providecommand{\BIBentryALTinterwordstretchfactor}{4}
\providecommand{\BIBentryALTinterwordspacing}{\spaceskip=\fontdimen2\font plus
\BIBentryALTinterwordstretchfactor\fontdimen3\font minus \fontdimen4\font\relax}
\providecommand{\BIBforeignlanguage}[2]{{%
\expandafter\ifx\csname l@#1\endcsname\relax
\typeout{** WARNING: IEEEtran.bst: No hyphenation pattern has been}%
\typeout{** loaded for the language `#1'. Using the pattern for}%
\typeout{** the default language instead.}%
\else
\language=\csname l@#1\endcsname
\fi
#2}}
\providecommand{\BIBdecl}{\relax}
\BIBdecl

\bibitem{bahl2023affordances}
S.~Bahl, R.~Mendonca \emph{et~al.}, ``Affordances from human videos as a versatile representation for robotics,'' in \emph{CVPR}, 2023, pp. 13\,778--13\,790.

\bibitem{ju2024robo}
Y.~Ju, K.~Hu \emph{et~al.}, ``Robo-abc: Affordance generalization beyond categories via semantic correspondence for robot manipulation,'' in \emph{ECCV}, 2024, pp. 222--239.

\bibitem{kuang2024ram}
Y.~Kuang, J.~Ye \emph{et~al.}, ``Ram: Retrieval-based affordance transfer for generalizable zero-shot robotic manipulation,'' \emph{arXiv preprint arXiv:2407.04689}, 2024.

\bibitem{dong2024rtagrasp}
W.~Dong, D.~Huang \emph{et~al.}, ``Rtagrasp: Learning task-oriented grasping from human videos via retrieval, transfer, and alignment,'' \emph{arXiv preprint arXiv:2409.16033}, 2024.

\bibitem{tang2023graspgpt}
C.~Tang, D.~Huang \emph{et~al.}, ``Graspgpt: Leveraging semantic knowledge from a large language model for task-oriented grasping,'' \emph{IEEE Robotics and Automation Letters}, 2023.

\bibitem{tang2024foundationgrasp}
C.~Tang, D.~Huang, W.~Dong, R.~Xu, and H.~Zhang, ``Foundationgrasp: Generalizable task-oriented grasping with foundation models,'' \emph{arXiv preprint arXiv:2404.10399}, 2024.

\bibitem{li2024ovgnet}
M.~Li, Q.~Zhao, S.~Lyu, C.~Wang, Y.~Ma, G.~Cheng, and C.~Yang, ``Ovgnet: A unified visual-linguistic framework for open-vocabulary robotic grasping,'' in \emph{IROS}, 2024, pp. 7507--7513.

\bibitem{zheng2024gaussiangrasper}
Y.~Zheng, X.~Chen, Y.~Zheng \emph{et~al.}, ``Gaussiangrasper: 3d language gaussian splatting for open-vocabulary robotic grasping,'' \emph{arXiv preprint arXiv:2403.09637}, 2024.

\bibitem{bhat2024hifi}
V.~Bhat, P.~Krishnamurthy \emph{et~al.}, ``Hifi-cs: Towards open vocabulary visual grounding for robotic grasping using vision-language models,'' \emph{arXiv preprint arXiv:2409.10419}, 2024.

\bibitem{tziafas2024towards}
G.~Tziafas and H.~Kasaei, ``Towards open-world grasping with large vision-language models,'' \emph{arXiv preprint arXiv:2406.18722}, 2024.

\bibitem{van2024open}
T.~van Oort, D.~Miller, W.~N. Browne, N.~Marticorena, J.~Haviland, and N.~Suenderhauf, ``Open-vocabulary part-based grasping,'' \emph{arXiv preprint arXiv:2406.05951}, 2024.

\bibitem{liu2024grounding}
S.~Liu, Z.~Zeng \emph{et~al.}, ``Grounding dino: Marrying dino with grounded pre-training for open-set object detection,'' in \emph{ECCV}, 2024, pp. 38--55.

\bibitem{sun2023going}
P.~Sun, S.~Chen \emph{et~al.}, ``Going denser with open-vocabulary part segmentation,'' in \emph{ICCV}, 2023, pp. 15\,453--15\,465.

\bibitem{mirjalili2024lan}
R.~Mirjalili, M.~Krawez, S.~Silenzi, Y.~Blei, and W.~Burgard, ``Lan-grasp: An effective approach to semantic object grasping using large language models,'' in \emph{First Workshop on Vision-Language Models for Navigation and Manipulation at ICRA 2024}.

\bibitem{tong2024oval}
E.~Tong, A.~Opipari, S.~Lewis, Z.~Zeng, and O.~C. Jenkins, ``Oval-prompt: Open-vocabulary affordance localization for robot manipulation through llm affordance-grounding,'' \emph{arXiv preprint arXiv:2404.11000}, 2024.

\bibitem{luo2022learning}
H.~Luo, W.~Zhai \emph{et~al.}, ``Learning affordance grounding from exocentric images,'' in \emph{CVPR}, 2022, pp. 2252--2261.

\bibitem{chen2023affordance}
J.~Chen, D.~Gao \emph{et~al.}, ``Affordance grounding from demonstration video to target image,'' in \emph{CVPR}, 2023, pp. 6799--6808.

\bibitem{li2024laso}
Y.~Li, N.~Zhao, J.~Xiao \emph{et~al.}, ``Laso: Language-guided affordance segmentation on 3d object,'' in \emph{CVPR}, 2024, pp. 14\,251--14\,260.

\bibitem{ma2024glover}
T.~Ma, Z.~Wang, J.~Zhou, M.~Wang, and J.~Liang, ``Glover: Generalizable open-vocabulary affordance reasoning for task-oriented grasping,'' \emph{arXiv preprint arXiv:2411.12286}, 2024.

\bibitem{murali2021same}
A.~Murali, W.~Liu \emph{et~al.}, ``Same object, different grasps: Data and semantic knowledge for task-oriented grasping,'' in \emph{CORL}, 2021, pp. 1540--1557.

\bibitem{song2023learning}
Y.~Song, P.~Sun \emph{et~al.}, ``Learning 6-dof fine-grained grasp detection based on part affordance grounding,'' \emph{arXiv preprint arXiv:2301.11564}, 2023.

\bibitem{lerftogo2023}
A.~Rashid, S.~Sharma \emph{et~al.}, ``Language embedded radiance fields for zero-shot task-oriented grasping,'' in \emph{7th Annual Conference on Robot Learning}, 2023.

\bibitem{kerr2023lerf}
J.~Kerr, C.~M. Kim \emph{et~al.}, ``Lerf: Language embedded radiance fields,'' in \emph{ICCV}, 2023, pp. 19\,729--19\,739.

\bibitem{caron2021emerging}
M.~Caron, H.~Touvron \emph{et~al.}, ``Emerging properties in self-supervised vision transformers,'' in \emph{ICCV}, 2021, pp. 9650--9660.

\bibitem{li2024shapegrasp}
S.~Li, S.~Bhagat, J.~Campbell, Y.~Xie, W.~Kim, K.~Sycara, and S.~Stepputtis, ``Shapegrasp: Zero-shot task-oriented grasping with large language models through geometric decomposition,'' \emph{arXiv preprint arXiv:2403.18062}, 2024.

\bibitem{jin2024reasoning}
S.~Jin, J.~Xu, Y.~Lei, and L.~Zhang, ``Reasoning grasping via multimodal large language model,'' \emph{arXiv preprint arXiv:2402.06798}, 2024.

\bibitem{xu2024rt}
J.~Xu, S.~Jin \emph{et~al.}, ``Rt-grasp: Reasoning tuning robotic grasping via multi-modal large language model,'' in \emph{IROS}, 2024, pp. 7323--7330.

\bibitem{qian2024thinkgrasp}
Y.~Qian, X.~Zhu \emph{et~al.}, ``Thinkgrasp: A vision-language system for strategic part grasping in clutter,'' \emph{arXiv preprint arXiv:2407.11298}, 2024.

\bibitem{hurst2024gpt}
A.~Hurst, A.~Lerer \emph{et~al.}, ``Gpt-4o system card,'' \emph{arXiv preprint arXiv:2410.21276}, 2024.

\bibitem{10167687}
H.-S. Fang, C.~Wang \emph{et~al.}, ``Anygrasp: Robust and efficient grasp perception in spatial and temporal domains,'' \emph{IEEE Transactions on Robotics}, pp. 3929--3945, 2023.

\bibitem{fang2020graspnet}
H.-S. Fang, C.~Wang, M.~Gou, and C.~Lu, ``Graspnet-1billion: A large-scale benchmark for general object grasping,'' in \emph{CVPR}, 2020, pp. 11\,444--11\,453.

\bibitem{coumans2016pybullet}
E.~Coumans and Y.~Bai, ``Pybullet, a python module for physics simulation for games, robotics and machine learning,'' 2016.

\end{thebibliography}

\end{document}